# Gaussian Process Networks

**Nir Friedman**
School of Computer Science & Engineering
Hebrew University
Jerusalem, 91904, ISRAEL
nir@cs.huji.ac.il

**Iftach Nachman**
Center for Neural Computation and
School of Computer Science & Engineering
Hebrew University
Jerusalem, 91904, ISRAEL
iftach@cs.huji.ac.il

## Abstract

In this paper we address the problem of learning the structure of a Bayesian network in domains with continuous variables. This task requires a procedure for comparing different candidate structures. In the Bayesian framework, this is done by evaluating the *marginal likelihood* of the data given a candidate structure. This term can be computed in closed-form for standard parametric families (e.g., Gaussians), and can be approximated, at some computational cost, for some semi-parametric families (e.g., mixtures of Gaussians).

We present a new family of continuous variable probabilistic networks that are based on *Gaussian Process* priors. These priors are semi-parametric in nature and can learn almost arbitrary noisy functional relations. Using these priors, we can directly compute marginal likelihoods for structure learning. The resulting method can discover a wide range of functional dependencies in multivariate data. We develop the Bayesian score of Gaussian Process Networks and describe how to learn them from data. We present empirical results on artificial data as well as on real-life domains with non-linear dependencies.

## 1 Introduction

Bayesian networks are a language for representing joint probability distributions of many random variables. They are particularly effective in domains where the interactions between variables are fairly local: each variable directly depends on a small set of other variables. Bayesian networks have been applied extensively for modeling complex domains. This success is due both to the flexibility of the models and to the naturalness of incorporating expert knowledge into the domain. An important ingredient for many applications is the ability to induce models from data. This ability allows to complement expert knowledge with data to improve performance of a system.

In the last decade there has been an active research effort to develop the theory and algorithms for learning of Bayesian networks from data. This includes methods for parameter learning [1, 19, 27] and structure learning [3, 6, 16, 26]. Using structure learning procedures we can learn about the structure of interactions between variables in an unknown domain.

Part of our motivation comes from an ongoing project that applies such structure learning methods to molecular biology problems [10]. This project attempts to understand transcription of genes: A gene is *expressed* via a process that *transcribes* it into an RNA sequence, and this RNA sequence is in turn *translated* into a protein molecule. Recent technical breakthroughs in molecular biology enable biologists to measure of the expression levels of thousands of genes in one experiment [7, 20, 32]. The data generated from these experiments consists of instances, each one of which has thousands of attributes. These data sets can help us understand how a gene's transcription is effected by various aspects of the cell's metabolism, including the expression levels of other genes. The challenge is to recover this biological knowledge from such experiments (see, e.g., [18]).

There are several problems in learning from such data. In particular, in this paper we examine the problems raised by the quantitative nature of these measurements. In theory we might think of a gene as being either in "activated" and "suppressed" modes. Experience with this data, however, shows that discretization of the data loses much of the information [10]. Thus, we seek methods that can directly represent and learn interactions among continuous variables.

Another problematic aspect of this type of data is the large number of attributes (genes) that are measured (i.e., thousands) and the relatively few samples (i.e., dozens). Thus, we seek methods that are statistically robust and can detect dependencies among many possible alternatives.

The best understood approach for modeling continuous distributions in Bayesian network learning is based on Gaussian distributions [11]. This form of continuous Bayesian network can be learned using exact Bayesian derivations quite efficiently. Unfortunately, the expressive power of Gaussian networks is limited. Formally, "pure" Gaussian networks can only learn *linear* dependencies among the measured variables.

This is a serious restriction when learning in domains with non-linear interactions, or domains where the nature of the interactions is unknown. A common way of avoiding this problem is to introduce hidden variables that represent mixtures of Gaussians (e.g., [28, 34]). An alternative approach that has been suggested is to learn with non-parametric densities [17].

In this paper we address the problem of learning continuous networks by using *Gaussian Process* priors. This class of priors is a flexible semi-parametric regression model. We call the networks learned using this method *Gaussian Process Networks*. The resulting learning algorithm is capable of learning a large range of dependencies from data.

This approach has several important properties. First, the Gaussian Process is a Bayesian method. Thus, the integration of this form of regression into the Bayesian framework of model selection is natural and fairly straightforward. This allows us to interpret the results of the learning as posterior probabilities, and to assess the posterior probability of various networks structures (e.g., using methods such as [9]). Second, the semi-parametric nature of the prior allows to learn many continuous functional dependencies. This is crucial for exploratory data analysis where there is little prior knowledge on the form of interactions we may encounter in data. In addition, the Gaussian Process is biased to find functional dependencies among the variables in the domain. This is a useful prior for domains where we believe there is a direct causal dependency between attributes.

In the remainder of this paper we review the Bayesian approach for learning Bayesian networks. We then review the definition of the Gaussian process prior in this setting and discuss how to combine the two to learn networks. Finally, we validate our approach on series of artificial examples that test its generalization capabilities and apply to few real-life data problems.

## 2 Learning Continuous Networks

### 2.1 Bayesian Structure Learning

Our goal is to learn Bayesian networks from fully observable data; i.e., we are given a data set $D = \{\mathbf{x}[1], \ldots, \mathbf{x}[M]\}$, where each $\mathbf{x}[j]$ is a complete assignment to the variables $X_1, \ldots, X_n$. The Bayesian learning paradigm requires us to specify a prior probability distribution $P(G)$ over the space of possible Bayesian network structures, and for each structure, a prior over the conditional probabilities. This prior is then updated using Bayesian conditioning to give a posterior distribution $P(G \mid D)$ over this space.

We start with the prior over structures, $P(G)$. Several priors have been proposed, all of which are quite simple. Without going into detail, a key property of all these priors is that they satisfy:

- **Structure modularity** The prior $P(G)$ can be written in the form

$$P(G) \propto \prod_i \rho(X_i, \mathrm{Pa}_G(X_i));$$

That is, the prior decomposes into a product, with a term for each family in $G$. In other words the choices of the families for the different nodes are independent a priori.

The next question we need to address is the prior over the conditional probability distributions (or density functions) $P(X_i \mid \mathrm{Pa}_G(X_i))$ needed for each network. Usually, these conditional distributions are represented in a parametric form. Thus, once we fix the parametric family, we can specify the conditional distribution by a vector of parameters $\theta_{X_i \mid \mathrm{Pa}_G(X_i)}$. The prior distribution over conditional distributions can now be phrased as a prior over values of the different $\theta_{X_i \mid \mathrm{Pa}_G(X_i)}$'s. For our purpose, we need only require that the prior satisfies two basic assumptions, as presented in [16]:

- **Parameter independence:** Let $\theta_{X_i \mid \mathrm{Pa}_G(X_i)}$ be the parameters specifying the behavior of the variable $X_i$ given the various instantiations to its parents $\mathbf{U}$. We require that

$$P(\theta_G \mid G) = \prod_i P(\theta_{X_i \mid \mathrm{Pa}_G(X_i)} \mid G) \quad (1)$$

That is, we assume that the parameters for different conditional distributions are a priori independent.

- **Parameter modularity:** Let $G$ and $G'$ be two graphs in which $\mathrm{Pa}_G(X_i) = \mathrm{Pa}_{G'}(X_i) = \mathbf{U}$ then

$$P(\theta_{X_i \mid \mathbf{U}} \mid G) = P(\theta_{X_i \mid \mathbf{U}} \mid G') \quad (2)$$

That is, the prior for a conditional distribution depends only on the choice of parents for $X_i$ and is independent of other aspects of the graph $G$.

Once we define the prior, we can examine the form of the posterior probability. Using Bayes rule, we have that

$$P(G \mid D) \propto P(D \mid G)P(G).$$

The term $P(D \mid G)$ is the *marginal probability* of the data given $G$ and is defined as the integration over all possible parameter values for $G$

$$P(D \mid G) = \int P(D \mid G, \theta_G) P(\theta_G \mid G) d\theta_G$$

Alternatively, we can define $P(D \mid G)$ using the chain rule:

$$P(D \mid G) = \prod_i P(\mathbf{x}[i] \mid \mathbf{x}[1], \ldots, \mathbf{x}[i-1], G)$$

where $P(\mathbf{x}[i] \mid \mathbf{x}[1], \ldots, \mathbf{x}[i-1], G)$ is the probability given to the $i$'th instance after observing the previous $i - 1$ instances.

Using the above assumptions, one can show (see [16]) that if $D$ is complete and the prior probability satisfies *parameter independence*, and *parameter modularity*, then

$$P(D \mid G) = \prod_i score(X_i, \mathrm{Pa}_G(X_i) \mid D).$$

where the conditional score $score(X_i, \mathbf{U} \mid D)$ is

$$P(x_i[1], \ldots, x_i[M] \mid \mathbf{u}[1], \ldots, \mathbf{u}[M], A[\mathrm{Pa}_G(X_i) = \mathbf{U}]).$$

That is, the probability assigned to the sequence of values $X_i$ given the observed values of $\mathbf{U}$ and the assumption that $X_i$'s parents are exactly $\mathbf{U}$.

If the prior $P(G)$ satisfies structure modularity, we can also conclude that the posterior probability decomposes:

$$P(G \mid D) \propto \prod_i \rho(X_i, \mathrm{Pa}_G(X_i)) score(X_i, \mathrm{Pa}_G(X_i) \mid D).$$

This decomposition of the score is crucial for learning structure. A *local* search procedure that changes one arc at each move can efficiently evaluate the gains made by adding or removing an arc. An example of such a procedure is a greedy hill-climbing procedure that at each step performs the local change that results in the maximal gain, until it reaches a local maximum. Although this procedure does not necessarily find a global maximum, it does perform well in practice. Examples of other local search procedures that advance in one-arc changes include beam-search, stochastic hill-climbing, and simulated annealing.

### 2.2 Continuous Variable Networks

To learn Bayesian networks we need to choose parametric families for representing and learning conditional densities.

There are several possible choices. We briefly mention these here.

The simplest and best understood families of conditional densities are the *linear Gaussian* models. In this model, if $\mathrm{Pa}_G(X) = \{U_1, \ldots, U_k\}$, we assume that

$$P(X \mid u_1, \ldots, u_k) \sim N(a_0 + \sum_i a_i \cdot u_i, \sigma^2).$$

That is, $X$ is normally distributed around a mean that depends *linearly* on the values of its parents. The variance of this normal distribution is independent of the parents' values. In this representation $\theta_{X \mid \{U_1, \ldots, U_k\}} = \langle a_0, \ldots, a_k, \sigma \rangle$.

Bayesian learning of such families is developed by Geiger and Heckerman [11, 15, 12]. While we do not go into details, we note that for this parametric family, the Bayesian score for each family can be computed exactly and quite efficiently.

The drawback of Gaussian networks is that their representation is limited to modeling linear dependencies between variables. Thus, if the dependencies in the data are significantly non-linear, the score of the parents choices can be misleading and thus result in a network that poorly reflects the dependencies in the data (and also performs poorly in predictions).

A possible approach to overcome the limitations of Gaussian models is to consider *mixtures of Gaussians* [8, 34]. In this approach we model the conditional distribution as a weighted mixture

$$P(X \mid \mathbf{U}) = \sum_j w_j f_j(X \mid \mathbf{U})$$

where each $f_j$ is a linear Gaussian distribution. In theory, such mixtures can approximate a wide range of conditional distributions. In particular, they can represent multi-modal distributions, and thus can represent relationships that are not purely functional.

Learning such mixture models, however, presents problems. Exact computation of the marginal likelihood of such a family cannot be done in closed form. Instead, we have to resort to approximations, such as the Laplace approximation [23, 4, 22]. This, in turn, requires us to find the MAP parameters given the data, which is a non-linear optimization problem in a space with many local maxima. Thus, in practice, we usually need nontrivial amount of data and running time to learn a mixture with moderate number of components.

An alternative approach which is non-Bayesian in nature was proposed by Hofmann and Tresp [17]. They use non-parametric kernel methods for predictions. Roughly speaking, given training examples $\mathbf{x}[1], \ldots, \mathbf{x}[M]$, the kernel estimate for $P(\mathbf{X})$ is

$$P_{kernel}(\mathbf{x}) = \frac{1}{M} \sum_{m=1}^{M} g\left(\frac{1}{\sigma} \|\mathbf{x} - \mathbf{x}[m]\|_2\right)$$

where $g()$ is a kernel function and $\sigma$ is a "smoothing" parameter. A common choice is to take $g$ to be the pdf of a normal distribution with zero mean and unit variance. Hofmann and Tresp use such estimates to find the conditional distribution by setting $P(x \mid \mathbf{u}) = P_{kernel}(x, u)/P_{kernel}(u)$.

Kernel methods are extremely flexible density estimators. Their performance depends crucially on the smoothness parameter. Thus, we need to tune this parameter to ensure that the data is not over-fit or over-smoothed. This is usually done by cross-validation testing. In particular, Hofmann and Tresp use a leave-one-out cross-validation procedure. In addition, we need to find a way of comparing the score of different network structures in this non-parametric setting. Hofmann and Tresp suggest to do so by comparing a cross-validated estimate of the logarithmic loss of each family. This is essentially an estimate of the out-of-sample loss the family will incur on new data. To summarize, for each family, Hofmann and Tresp's procedure searches for the parameters that minimize the log-loss in cross validation estimate, and then return this log-loss estimate as the score of the family.

## 3 Gaussian Process priors

In recent years, there has been much interest in the use of Gaussian Process priors for regression [33] as well as for classification [13]. It can be shown that predictors like feed-forward neural networks and radial-basis function networks converge to Gaussian process predictors as the number of internal nodes goes to infinity [21]. We now review the basics of the Gaussian Process prior and its use in regression.

Consider a set of variables $\mathbf{U}$. We want to model a prior over a variable $X$ which we believe to be a function of $\mathbf{U}$. We can treat the value of $X$ for each value $\mathbf{u}$ as a random variable. More formally, a *stochastic process* over $\mathbf{U}$ is a function that assigns to each $\mathbf{u} \in val(\mathbf{U})$ a random variable $X(\mathbf{u})$. The process is said to be a *Gaussian process* (GP) if for each finite set of values, $\mathbf{u}_{1:M} = \{\mathbf{u}[1], \ldots, \mathbf{u}[M]\}$, the distribution over the corresponding random variables $\mathbf{x}_{1:M} = \{X[1], \ldots, X[M]\}$ (where $X[m] = X(\mathbf{u}[m])$) is a multivariate normal distribution.

To specify such a process, we need a way of describing the mean value of each variable $X(\mathbf{u})$ and the co-variance matrix for each finite subset of values we chose. This is done, by specifying two functions:

- A mean function $\mu(\mathbf{u})$, so that $E[X(\mathbf{u})] = \mu(\mathbf{u})$.
- A covariance function $C(\mathbf{u}, \mathbf{u}')$, so that $Cov[X(\mathbf{u}), X(\mathbf{u}')] = C(\mathbf{u}, \mathbf{u}')$.

The joint distribution of $\mathbf{x}_{1:M}$ is therefore:

$$P(\mathbf{x}_{1:M} | \mathbf{u}_{1:M}) \qquad (3)$$
$$= \frac{1}{Z} \exp\left(-\frac{1}{2}(\mathbf{x}_{1:M} - \mu_{1:M})^T C_{1:M}^{-1} (\mathbf{x}_{1:M} - \mu_{1:M})\right)$$

where $\mu_{1:M}$ is the vector of means $\langle \mu(\mathbf{u}[1]), \ldots, \mu(\mathbf{u}[M]) \rangle$ and $C_{1:M}$ is the covariance matrix with the $(i, j)$ entry $C(\mathbf{u}[i], \mathbf{u}[j])$.

### 3.1 Prediction

Before we discuss the covariance function $C$ and its parameters, let us see how we use the GP to predict the value of the process at a new point. We shall assume $\mu(\mathbf{u}) = 0$ from now on.

Assume we already observed $M$ points $\mathbf{x}_{1:M}$ given $\mathbf{u}_{1:M}$, and we are given a parametrized covariance function. By the definition of the Gaussian process $P(\mathbf{X}_{1:M}, X_{M+1} | \mathbf{U}_{1:M}, \mathbf{U}_{M+1})$ is a $M + 1$-dimensional Gaussian distribution. We since we observed the values $\mathbf{X}_{1:M}$, we compute the conditional distribution over $X_{M+1}$ given these observation. A basic property of multivariate Gaussian distributions is that the conditional distribution given the value of some of the variables is also a Gaussian distribution. Thus, the conditional distribution $P(X_{M+1} | \mathbf{X}_{1:M}, \mathbf{U}_{1:M}, \mathbf{U}_{M+1})$ is a univariate Gaussian distribution. Using properties of Gaussian distributions we compute define the mean and variance of this distribution using:

$$\mu_{M+1} = \mathbf{k}^T C_M^{-1} \mathbf{x}_{1:M} \qquad (4)$$
$$\sigma_{M+1} = \kappa - \mathbf{k}^T C_M^{-1} \mathbf{k} \qquad (5)$$

Where $\mathbf{k} = (C(\mathbf{u}[M+1], \mathbf{u}[1]), \ldots, C(\mathbf{u}[M+1], \mathbf{u}[M]))$ and $\kappa = C(\mathbf{u}[M+1], \mathbf{u}[M+1])$. In other words, having observed $M$ values of the process we can represent the conditional density at any new coordinate $x$ using $C_M$, the covariance matrix calculated for the first $M$ points.

### 3.2 Covariance Functions

We now deal with the issue of covariance functions. As we can see, this function determines the prior over functions. The only constraint on the covariance is that it should produce positive semidefinite matrices.

In general, if the covariance of two close points is large, then the prior prefers smooth functions. The covariance between points further away determines properties like periodicity, smoothness, and amplitude of the learned functions. These aspects of the covariance functions are controlled by its *hyperparameters* $\theta$. For example, Williams and Rasmussen [33] suggest the following function:

$$C(\mathbf{u}, \mathbf{u}' : \theta) = \theta_0 \exp\left\{-\frac{1}{2} \sum_{U_k \in \mathbf{U}} \frac{(u_k - u_k')^2}{\lambda_k^2}\right\}$$
$$+ \theta_1 + \theta_2 \sum_{U_k \in \mathbf{U}} u_k u_k' + \theta_3 \delta_{\mathbf{u}, \mathbf{u}'} \qquad (6)$$

In this function each hyperparameter controls a different characteristic of the learned functions. The hyperparameter $\theta_0$ controls the amplitude of variation of the function. The hyperparameter $\theta_1$ controls how far can the whole function be shifted from the zero line. The hyperparameter $\theta_2$ accounts for linear tendencies in the function. And the hyperparameter $\theta_3$ is the variance of an uncorrelated white noise, which is added on top of the function. The hyperparameters $\lambda_k$ are the length scales of the different directions in $\mathbf{u}$, over which the function changes considerably.

What value of hyperparameters should we use in $C$ when constructing the Gaussian Process density ? The Bayesian approach is to assign the hyperparameters a prior, and then integrate over them. Let $D = \{\mathbf{u}_{1:M}, \mathbf{x}_{1:M}\}$, then we should make predictions as

$$P(x[M+1]|\mathbf{u}[M+1], D)$$
$$= \int P(x[M+1]|\mathbf{u}[M+1], D, \theta) P(\theta|D) d\theta$$

As this integral is usually intractable, we can try to approximate it. One way is to use $\tilde{\theta}$, the *maximum a posteriori* estimator for $\theta$, as suggested in [13]. Another option is performing a numerical integration using a Monte Carlo method (as in [33]).

## 4 Learning Networks with Gaussian Process priors

We now examine Networks with Gaussian Process priors. As stated above, we make the parameter independence and modularity assumptions. Thus, to define the prior we need to evaluate the score of a variable $X$ given a set $\mathbf{U}$ of parent variables.

Recall that the *score*$(X, \mathbf{U} \mid D)$ is defined

$$P(x[1], \ldots, x[M] \mid \mathbf{u}[1], \ldots, \mathbf{u}[M], A[\text{Pa}_G(X_i) = \mathbf{U}]).$$

To define these terms, we need to define a Gaussian Process prior for $X$ as a function of $\mathbf{U}$. As before, we will assume that the mean function is 0, and thus we only need to choose a covariance function $C_{X|\mathbf{U}}$.

Once we choose such a covariance function, the score is easy to compute. The Gaussian Process prior implies that $x[1], \ldots, x[M]$ are normally distributed with the covariance function specified by $\mathbf{U}[1], \ldots, \mathbf{U}[M]$. Thus, the score with respect to the covariance function $C(\cdot, \cdot : \theta)$ is

$$score(X_i, \mathbf{U} \mid D, \theta)$$
$$= (2\pi)^{-\frac{M}{2}} |C_\mathbf{U}|^{-\frac{1}{2}} \exp\left(-\frac{1}{2} \mathbf{x}_{1:M}^T C_\mathbf{U}^{-1} \mathbf{x}_{1:M}\right),$$

where $C_\mathbf{U}$ is the covariance matrix defined by the covariance function $C()$ and the values $\mathbf{u}_1, \ldots, \mathbf{u}_M$.

We see that given a Gaussian Process prior, the computation of marginal probability can be done is closed form. In this sense, the Gaussian Process prior is very appealing. It can learn a wide range of functional dependencies, and we can compute the Bayesian score exactly. In this sense, the Gaussian Process priors fit well with the Bayesian model selection approach of learning Bayesian network structure.

In practice, we usually do not fix the parametrized covariance function in advance. Instead, we select a family of priors, such as the ones of (6). Thus, the proper score would require us to integrate over the hyperparameters:

$$score(X, \mathbf{U} \mid D) = \int score(X, \mathbf{U} \mid D, \theta) P(\theta) d\theta.$$

Unfortunately, we do not know how to perform this integration in closed form, since $score(X, \mathbf{U} \mid D, \theta)$ is a complex function of $\theta$.

The approach we take, which is quite common in many other applications of Bayesian methods, is to approximate this integral with the MAP hyperparameters. Thus, we approximate

$$score(X, \mathbf{U} \mid D) \approx \max_\theta score(X, \mathbf{U} \mid D, \theta) P(\theta).$$

This approximation is reasonable if the posterior probability over hyperparameters is sharply peaked over a single maximum. In such situations, most of the integral is determined by the area near the MAP parameters. A slightly better approximation is the Laplace approximation, where the posterior probability in the integral is approximated as a Gaussian distribution over the parameters $\theta$ (see, e.g. [5]). This however requires the calculation of the Hessian of the log posterior probability, which can be time consuming. We therefore use an estimate for this term, which scales like $\frac{K}{2} log(N)$, where $K$ in our case is the number of hyperparameters of the covariance function. The resulting estimate is in the spirit of the *Bayesian information criterion (BIC)* of Schwarz [25], and the MDL score of Rissanen [24], having a term which penalizes the model for over-complexity.

To score a family $\mathbf{X}$ given $\mathbf{U}$, we perform conjugate gradient ascent to search for the MAP parameters. The evaluation of each point during the search requires to invert and to compute the determinant of an $M$ by $M$ matrix. Thus, the computational costs of this closed form equation is $O(M^3)$ in naive implementations. This operation is repeated in each iteration of the hyperparameter optimization step. In practice this optimization converges quite rapidly (10-20 iterations).

## 5 Experimental Evaluation

We first want to test the GP score on the simplest case. We therefore ask the following question: given two variables, $X$ and $Y$, with some noisy functional dependence between them, will the GPN learner prefer the network where X is independent of Y, or the one in which they are dependent. Furthermore, we expect that, up to a certain noise level, the GP learner will prefer the direction for which it can fit a "nice" function, since such a function is more likely in a GP prior. For example, in Figure 1 we see a noisy quadratic dependence. The GP prior will assign a very low likelihood to the $X \to Y$ dependence, since it is hard to fit a function in this direction, while the dependence $Y \to X$ will get a higher probability, as it can be explained by a quadratic functional dependence with a certain noise width at each point.

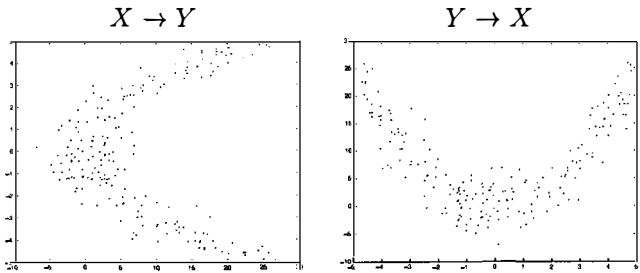

Figure 1: An example of a non-invertible dependence between X and Y. The explanation $X \to Y$ does not have a functional form, whereas $Y \to X$ can be explained as a noisy function.

To test this, we produced data sets of two variables with dependencies of linear, quadratic (as in Figure 1), cubic and sinusoidal nature. On top of the functional dependence, a non-correlated Gaussian noise was added. For each case we compared between the different network models, in terms of the GP network scores for the training set, and the log likelihood of the test set when that particular model was used for prediction. This was done for different noise levels, and different training set sizes.

Figures 3 and 2 show the dependence of those measures on the function noise level. We observe that for the true dependency model, the prediction quality and the GP family score rise as the level of noise drops. We see that even for noise levels as high as 1.5 times the dependent variable amplitude, the true dependence is still preferred over the no-dependence model. We also see that the direction of dependence is clear in the non-invertible cases, like the sinusoidal and the quadratic dependencies. In those cases, the score of the "wrong" direction dependency is as low as the no-dependency model. The cubic data set in our case is borderline-invertible, and so the distinction is less clear cut. For the linear case, if the slope is not too steep, both directions have a functional form, and so no one direction is preferred over the other. The Gaussian Process preference for functional direction can be useful when learning causal networks if we assume the interactions in our domain are functional.

We next compare the GP network learning method against the two continuous variable models described in Section 2.2: the Gaussian network model (with the BGe scoring metric [11]), and the kernel network. We start with two variable networks, with the same four types of functional relations as described before. Figure 4 shows the prediction quality of the three methods on those data sets, com-paring the log loss of the predictions made by the dependent model to those made by the independent model. One can see that for the quadratic and sinusoidal relations, both far from linear,, the Gaussian method prediction quality is the same for both models, while the GP learner continually performs better with the dependent model. The kernel method, which is insensitive to directionality or linearity, also performs better with the dependent model.

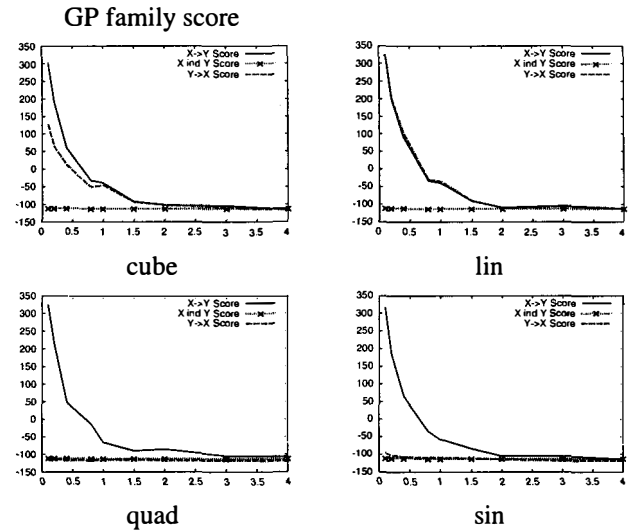

Figure 2: GP family scores as a function of sample noise, for different functional dependencies. Plots are shown for 3 network models: the no-dependency network, the $X \to Y$ "true" network and the $Y \to X$ "opposite direction" network. The shown functional dependencies of Y on X are linear, quadratic, cubic, and sinusoidal. The $Y$ axis is the GP score for each family, and the $X$ axis is the sample noise units in standard deviations of the dependent variable $(Y)$.

We now turn to comparing the reconstruction capability of the three methods. We start with small artificial networks with different functional relations, and check which method reconstructs the true network with higher accuracy. We sampled 50 and 100 instance data sets from 3 variable networks of all possible architectures, with linear, quadratic, sinusoidal or mixed functional relations. A non-correlated noise of width 0.4 of the variable's amplitude was added. We applied the three network learning methods on these data sets. Both GP and kernel methods performed well in reconstructing the true PDAG of the generating network, with the GP performing only slightly better. However, the GP does significantly better in identifying the original DAG for data sets with non-invertible connections (quadratic and sinusoidal). In those cases, as expected, the GP learner orients the arcs in the "true" functional direction, while the kernel method does not necessarily do so. The Gaussian network model does not perform as well in this task, where in most of the cases with non-linear con-nections, the learned networks are missing some of the arcs. This is no surprise, since the best linear-Gaussian model one can fit to a non-linear function often has a large variance, making this connection low scoring.

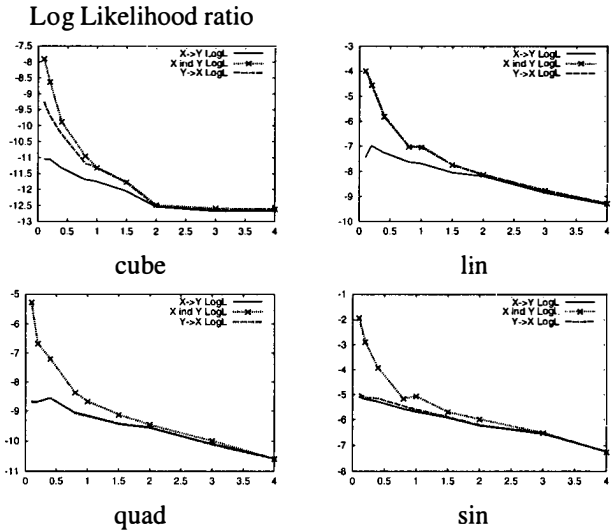

Figure 3: Prediction accuracy as a function of sample noise, for the same sets of Figure 2. The $Y$ axis is the average log loss of a test data set following parameter optimization on a different set. The $X$ axis is the sample noise in standard deviations of the dependent variable.

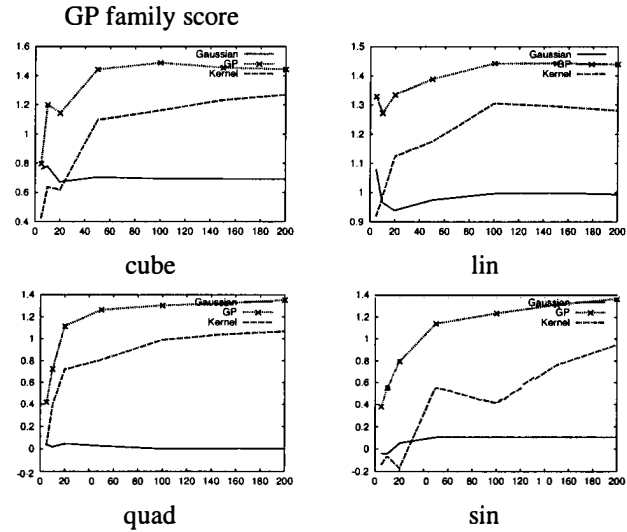

Figure 4: Sample complexity comparison for the Gaussian, Gaussian Process and Kernel methods. The plots show log likelihood ratio of the test set, between the no-dependency model and the $X \to Y$ model. Four different functional relations between $X$ and $Y$ were tested. Both training and test set have a noise level of 0.4 standard deviation of the dependent variable.

### 5.1 Real life data

We next wanted to test Gaussian Process Networks on real world data sets of continuous attributes, comparing it to the other two methods. We use three data sets from the UCI machine learning repository [2]. These data sets are:

- **Boston housing data set** - a data set describing different aspects of neighborhoods in the Boston area, and the median price of houses in those neighborhoods. The data set contains 506 samples with 14 attributes. 300 samples were used as a test set.
- **Abalone data set** - a data set of physical measurements of abalones. The data set contains 4177 samples with 9 attributes. 300 samples were used as a test set.
- **Glass identification data set** - a data describing the material concentrations in glasses, with a class attribute denoting the type of the glass. The data set contains 214 samples with 10 attributes. 64 samples were used as a test set.

For each data set, we performed structure learning with each method, using subsets of the original data set, which was permuted in a random order. We then used the learned structure and the optimized parameters to predict the likelihood of the corresponding test sets, which were not included in the training sets. Some of the attributes in those data sets are either discrete (such as class attributes), or have only few values in the data. To accommodate these variables, we used the hybrid approach as described, for example, in [14]. In this approach, all discrete variables are forced to precede all continuous variables. For each continuous variable $X$ having some mixture of continuous parents $\mathbf{U}_c$ and discrete parents $\mathbf{U}_d$, we model the distribution $P(X \mid U_c)$ separately for each state of $U_d$. The score for such a family is given by

$$score(X, \mathbf{U}_c, \mathbf{U}_d \mid D) = \sum_{\mathbf{u}_d \in \mathcal{U}_d} score(X, \mathbf{U}_c \mid D_{\mathbf{u}_d})$$

where $\mathcal{U}_d$ is the set of values taken by $\mathbf{U}_d$, and $D_{U_d}$ is the subset of data where $\mathbf{U}_d$ have values $\mathbf{u}_d$.

Table 1 lists the average log likelihood of the test set, for each method and each training set size. We note that both in the glass and abalone domains the Gaussian process method performs well in comparison to the Gaussian model, while the kernel method does not do as well. On the Boston domain, however, the kernel method seems to rate quite high. This is due to one variable (index of accessibility to radial highways) which only has nine values appearing in the data set. The kernel method assigns no parents to this variable, and learns a distribution composed of sharp "delta" peaks around those values. In cases like this, the kernel method has to be bounded not to learn distributions which are too sharp. Another option is to treat those vari-ables as discrete. In general, however, the Gaussian and GP methods, modeling only function-like relations, can not account for multi modal distributions.

Table 1: Average Log Loss on an independent test set achieved by the three methods for different training set sizes.

| Size | Boston | | | Abalone | | | Glass | | |
|---|---|---|---|---|---|---|---|---|---|
| | Gaussian | GP | Kernel | Gaussian | GP | Kernel | Gaussian | GP | Kernel |
| 10 | -53.78 | -28105.00 | -56.24 | -322.39 | -319.84 | -410.47 | -43.81 | -153.77 | -72.76 |
| 20 | -40.92 | -447.85 | -40.65 | 0.57 | -0.12 | -9.28 | -10.40 | -74.44 | -52.82 |
| 50 | -37.10 | -44.68 | -47.71 | 4.55 | 10.34 | -8.06 | -6.61 | -51.34 | -84.01 |
| 100 | -34.44 | -50.75 | -132.27 | 7.56 | 11.46 | -7.01 | -3.27 | -52.93 | -35.42 |
| 150 | -32.27 | -70.35 | 4.37 | 9.27 | 13.07 | -6.48 | -2.47 | -2.02 | -42.80 |
| 200 | -30.97 | -43.52 | 8.32 | 10.48 | 13.10 | -34.10 | | | |
| 300 | | | | 12.03 | 12.95 | -5.34 | | | |

Figure 5 shows two examples from the abalone domain of connections learned by the Gaussian Process network, plotted with the training samples. The GP learner clearly fits a non-linear function to the data, whose width varies according to the density of points in each area. Beyond the range of sample points, the width of the predicted function rises, as the uncertainty increases. The figure on the right is an example where the dependent variable is semi-discrete (number of rings), showing that the method is capable of handling this type of data as well. These examples show that the Gaussian Process Network methods can reveal interesting relations even under noisy measurements.

## 6 Discussion

In this paper we introduced the notion of Gaussian Process networks and developed the Bayesian score for learning these. We report on preliminary results that show that this method generalizes well from noisy data. The combination of this powerful regression technique with the flexible language of Bayesian networks seems like a promising tool for exploratory data analysis, causal structure discovery, prediction, and Bayesian classification.

There are several methods closely related to Gaussian Processes that are relevant to this work. Wahba [31, 29] makes a connection between Gaussian processes and reproducing kernel Hilbert spaces (RKHS), showing that the solution to the posterior Bayesian estimate of the Gaussian process (as in Equation 4) is also the solution to a spline smoothing problem posed as a variational minimization problem in an RKHS. The smoothing parameter is optimized using cross validation methods, whereas in the case of Gaussian process priors, we use a MAP estimate for the hyperparameters. In related works (e.g. [30]), the relevance of the different components of the function is estimated from the learned smoothing parameters. In Gaussian process methods there is a similar notion, judging the relevance of different input dimensions by their estimated lengthscales in the covariance function [21]. Inputs with estimated large lengthscales are deemed less relevant, because the function hardly changes in those directions. A promising direction for future research is guiding the search in the network space by those learned lengthscales, resulting in a more efficient and accurate search procedure. This is important when using the Gaussian process score method, as its computation is costly.

We are currently applying the Gaussian process network method to analyze biological time series data. Our hope is that by learning DBNs and the influences within them, we would be able to understand the structure of the dynamics that controls the generating processes. For example, we might learn that $dX$ depends on $Y$, which would give us a clue as to the effects of $Y$'s presence on $X$.

### Acknowledgements

This work was supported by Israel Science Foundation grant number 224/99-1 and by the generosity of the Michael Sacher fund. Nir Friedman was also supported by Harry & Abe Sherman Senior Lectureship in Computer Science. Iftach Nachman was also supported by the Center for Neural Computation, Hebrew University. Experiments reported here were run on equipment funded by an ISF Basic Equipment Grant.

## References

[1] J. Binder, D. Koller, S. Russell, and K. Kanazawa. Adaptive probabilistic networks with hidden variables. *Machine Learning*, 29:213–244, 1997.

[2] C.L. Blake and C.J. Merz. UCI repository of machine learning databases, 1998.

[3] W. Buntine. Theory refinement on Bayesian networks. In *UAI*, 1991.

[4] D. M. Chickering and D. Heckerman. Efficient approximations for the marginal likelihood of bayesian networks with hidden variables. *Machine Learning*, 29:181–212, 1997.

[5] D. M. Chickering and D. Heckerman. Efficient approximations for the marginal likelihood of bayesian networks with hidden variables. *Machine Learning*, 29:181–212, 1997.

[6] G. F. Cooper and E. Herskovits. A Bayesian method for the induction of probabilistic networks from data. *Machine Learning*, 9:309–347, 1992.

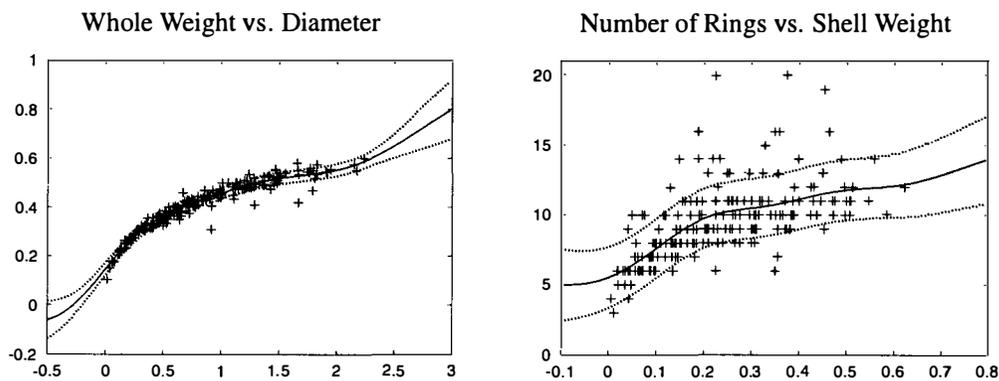

Figure 5: Two examples of function predictions made by the GP learner on the abalone data set. The values of the two attributes in the data set are shown as scattered points. The solid lines show the predicted mean function and its width at each point.

[7] J. DeRisi, V. Iyer, and P. Brown. Exploring the metabolic genetic control of gene expression on a genomic scale. *Science*, 278:680–686, 1997.

[8] N. Friedman, M. Goldszmidt, and T.J. Lee. Bayesian network classification with continuous attributes: Getting the best of both discretization and parametric fitting. In *ICML*, 1998.

[9] N. Friedman and D. Koller. Being Bayesian about Network Structure. In *UAI*, 2000.

[10] N. Friedman, M. Linial, I. Nachman, and D. Pe'er. Using bayesian networks to analyze expression data. In *RECOMB*, 2000.

[11] D. Geiger and D. Heckerman. Learning gaussian networks. MSR-TR-94-10, Microsoft Research, 1994.

[12] D. Geiger and D. Heckerman. Parameter priors for directed acyclic graphical models and characterization of several probability distributions. In *UAI*, 1999.

[13] M. N. Gibbs and D. J. C. MacKay. Variational Gaussian process classifiers. Unpublished manuscripts, available at http://wol.ra.phy.cam.ac.uk/mackay, 1997.

[14] D. Heckerman and D. Geiger. Learning Bayesian networks: a unification for discrete and Gaussian domains. In *UAI*, 1995.

[15] D. Heckerman and D. Geiger. Learning bayesian netwroks: A unification for discrete and gaussian domains. In *UAI*, 1995.

[16] D. Heckerman, D. Geiger, and D. M. Chickering. Learning Bayesian networks: The combination of knowledge and statistical data. *Machine Learning*, 20:197–243, 1995.

[17] R. Hofmann and V. Tresp. Discovering structure in continuous variables using bayesian networks. In *NIPS*, 1996.

[18] E. Lander. Array of hope. *Nature Genetics*, 21(1):3–4, January 1999.

[19] S. L. Lauritzen. The EM algorithm for graphical association models with missing data. *Computational Statistics and Data Analysis*, 19:191–201, 1995.

[20] D. J. Lockhart, H. Dong, M. C. Byrne, M. T. Follettie, M. V. Gallo, M. S. Chee, M. Mittmann, C. Want, M. Kobayashi, H. Horton, and E. L. Brown. DNA expression monitoring by hybridization of high density oligonucleotide arrays. *Nature Biotechnology*, 14:1675–1680, 1996.

[21] D. J. C. MacKay. Introduction to Gaussian processes. Unpublished manuscripts, available at http://wol.ra.phy.cam.ac.uk/mackay, 1998.

[22] Christopher Meek and David Heckerman. Structure and parameter learning for causal independence and causal interaction models. In *UAI* 1997.

[23] Cheeseman P. and Stutz J. Bayesian classification (AutoClass): Theory and results. In Fayyad U., Piatesky-Shapiro G., Smyth P., and Uthuruasmy R., editors, *Advances in Knowledge Discovery and Data Mining*, 1995.

[24] J. Rissanen. *Stochastic Complexity in Statistical Inquiry.* 1989.

[25] G. Schwarz. Estimating the dimension of a model. *Annals of Statistics*, 6:461–464, 1978.

[26] D. J. Spiegelhalter, A. P. Dawid, S. L. Lauritzen, and R. G. Cowell. Bayesian analysis in expert systems. *Statistical Science*, 8:219–283, 1993.

[27] D. J. Spiegelhalter and S. L. Lauritzen. Sequential updating of conditional probabilities on directed graphical structures. *Networks*, 20:579–605, 1990.

[28] B. Thiesson, C. Meek, D. M. Chickering, and D. Heckerman. Learning mixtures of Bayesian networks. In *UAI*, 1998.

[29] G. Wahba. *Spline Models for Observational Data.* Society for Industrial and Applied Mathematics, 1990.

[30] G. Wahba. Multivariate model building with additive, interaction, and tensor product thin plate splines. pages 491–504, 1991.

[31] G. Wahba. An introduction to model building with reproducing kernel hilbert spaces. Technical Report TR 1020, Univ. of Wisconsin, 2000.

[32] X. Wen, S. Furhmann, G. S. Micheals, D. B. Carr, S. Smith, J. L. Barker, and R. Somogyi. Large-scale temporal gene expression mapping of central nervous system development. *PNAS*, 95:334–339, 1998.

[33] C. K. I. Williams and C. E. Rasmussen. Gaussian processes for regression. In *NIPS*. 1996.

[34] L. Xu and M.I. Jordan. On convergence properties of the EM algorithm for Gaussian mixtures. *Neural Computation*, 8:129–151, 1996.